# A New Look at Causal Independence

David Heckerman    John S. Breese

Microsoft Research
One Microsoft Way
Redmond, WA 98052-6399
<heckerma—breese@microsoft.com>

## Abstract

Heckerman (1993) defined causal independence in terms of a set of temporal conditional independence statements. These statements formalized certain types of causal interaction where (1) the effect is independent of the order that causes are introduced and (2) the impact of a single cause on the effect does not depend on what other causes have previously been applied. In this paper, we introduce an equivalent atemporal characterization of causal independence based on a functional representation of the relationship between causes and the effect. In this representation, the interaction between causes and effect can be written as a nested decomposition of functions. Causal independence can be exploited by representing this decomposition in the belief network, resulting in representations that are more efficient for inference than general causal models. We present empirical results showing the benefits of a causal-independence representation for belief-network inference.

## 1 Introduction

Belief networks are often used as a modeling tool when there is uncertainty in the interaction between a set of causes and effects. A typical interaction between several causes and a single effect can be modeled with the belief network shown in Figure 1. In the figure, the variable $e$ represents an effect and the variables $c_1, \ldots, c_n$ represent $n$ causes of that effect. For binary discrete variables, this representation requires $2^n$ independent parameters to be specified. Consequently, the representation imposes intractable demands on both knowledge acquisition and inference.

In response to the intractability of knowledge acquisition, prototypical interactions such as the noisy-or model [?,?] have been developed. These models allow one to specify $n$ parameters to generate the conditional probability table for a variable $e$ as shown in Figure 1. Because the full table is used to characterize the relationship in the belief network, however, inference remains intractable.

figure=multfaul.eps,width=2in

Figure 1: A belief network for multiple causes and a single effect.

Last year at this conference, Heckerman defined causal independence in terms of temporal conditional-independence constraints on a set of variables [?]. These statements formalized certain types of causal interaction where (1) the effect is independent of the order that causes are introduced, and (2) the impact of a single cause on the effect does not depend on what other causes have previously been applied. The previous paper demonstrated how this definition generalizes the notion of a noisy-or and noisy-adder model, and indicated how a belief network representation of causal independence can be used to increase the speed of inference.

In this paper, we transform the previous temporal definition into an equivalent atemporal representation. In doing so, we find that causal independence is a special case of a generalization of the noisy-or developed by Srinivas (1993). It also allows us to define several classes of interaction models in terms of expressiveness and efficiency. Finally, we present some empirical results regarding the storage and inference savings associated with application of causal independence to real-world networks.

## 2 Temporal Definition of Causal Independence

In the temporal definition of causal independence, we associate a set of variables indexed by time with each cause and with the effect. We use $c_{jt}$ to denote the variable associated with cause $c_j$ at time $t$, and $e_t$ to denote the variable associated with the effect at time $t$. For all times $t$ and $t'$, we require the variables $c_{jt}$ and $c_{jt'}$ to have the same set of (possibly infinite) states.

figure=tempci.eps,width=3in

Figure 2: A temporal belief-network representation of causal independence.

figure=tworepa.eps,width=3in (a)
figure=tworepb.eps,width=3in (b)

Figure 3: Two representations of causal independence that are equivalent to the representation in Figure 2.

Under these conditions, we say that $c_1, \ldots, c_n$ are *causally independent with respect to* $e$ if the set of conditional-independent assertions

$$\forall t < t', c_j \ (e_{t'} \perp c_{1t}, \ldots, c_{j-1,t}, c_{j+1,t}, \ldots, c_{nt} \mid e_t, c_{jt}, c_{jt'}, c_{kt} = c_{kt'} \text{ for } k \neq j) \tag{1}$$

hold, where $(X \perp Y | Z)$ denotes the conditional-independence assertion "the sets of variables $X$ and $Y$ are independent, given $Z$." Note that Assertion 1 is somewhat unusual, in that independence is conditioned, in part, on the knowledge that the states of variables are equal, but otherwise undetermined ($c_{kt} = c_{kt'}$ for $k \neq j$). Assertion 1 states that if cause $c_j$ makes a transition from one state to another between $t$ and $t'$, and if no other cause makes a transition during this time interval, then the probability distribution over the effect at time $t'$ depends only on the state of the effect at time $t$ and on the transition made by $c_j$; the distribution does not depend on the other causation variables. Note that time is treated as an ordinal quantity in this definition and there is no need to have a continuous or discrete-interval model of time.

We can derive a belief-network representation of causal independence from this definition. First, for each cause, designate some state of its associated variables to be *distinguished.* For most real-world models, this state will be the one that has no bearing on the effect—that is, the "off" state—but we do not require this association. Second, let $\sigma$ be an ordering of the variables $\{c_1, \ldots, c_n\}$—we use $c_{\sigma i}$ to denote the $i$th variable in the ordering. Construct a belief network consisting of nodes $c_{\sigma 1}, \ldots, c_{\sigma n}$, and $e_0, e_{\sigma 1} \ldots, e_{\sigma n}$, as shown in Figure 2. In this belief network, node $e_0$ represents the effect when all causes take on their distinguished state. Node $c_{\sigma 1}$ represents the state of cause $c_{\sigma 1}$ after it has made a transition from its distinguished state (a transition may be the trivial transition, wherein the cause maintains its distinguished state). Node $e_{\sigma 1}$ represents the effect after only $c_{\sigma 1}$ has made the transition. In general, node $c_{\sigma i}$ represents the state of cause $c_{\sigma i}$ after it has made a (possibly trivial) transition from its distinguished state. Node $e_{\sigma i}$ represents the effect after causes $c_{\sigma 1}, \ldots, c_{\sigma i}$ have made their transitions. In particular, node $e_{\sigma n}$ represents the effect after all causes have made transitions. Thus, node $e_{\sigma n}$ corresponds to node $e$ in Figure 1.

The conditional independencies represented in the belief network of Figure 2 follow from the definition of causal independence. Conversely, given $n!$ belief networks of the form in Figure 2—one network for each possible ordering of the $n$ variables in the domain—we obtain the temporal definition of causal independence.

In terms of the number of parameter assessment for models of discrete-valued variables, causal independence yields a significant economy. As noted previously, the general multiple–cause interaction illustrated in Figure 1 requires $2^n$ separate assessments for binary variables: one parameter for each combination of the states of the parents. In contrast, the causal independence interaction illustrated in Figure 2 requires only $4n + 1$ assessments: four parameters for each node $e_{\sigma i}$ plus a single parameter for $e_0$. In Section 5 we discuss additional issues related to assessment of causal independence models.

## 3 An Atemporal Representation of Causal Independence

In this section, we transform the temporal definition into an atemporal form. The transformation is based on the observation that we can represent the belief network in Figure 2 as the belief network shown in Figure 3a. The double ovals represent *deterministic nodes*—nodes whose values are a deterministic function of their parents. Each node $\epsilon_{\sigma i}$ is a dummy node that encodes the uncertainty in the relationships among $e_{\sigma i}$ and its parents as described in Druzdel and Simon (1993) or, alternatively, Heckerman and Shachter (1994). We can think of the node $\epsilon_{\sigma i}$ as representing the causal mechanism that mediates the interaction between the parents of $e_{\sigma i}$ and $e_{\sigma i}$ itself [?], although we do not require this interpretation here. The definition presented in this section is atemporal in that it relies only on specification of a functional decomposition of the interaction, with no explicit representation of time.

Let variable $e'_{\sigma i}$ represent $e$ when all $c_j \neq c_{\sigma i}$ take on their distinguished variables. So, for example, $e'_{\sigma 1} = e_{\sigma 1}$. As we show in the following theorem, it turns out that if the relationships in Figure 3a are true for all orderings $\sigma$, then the relationships in Figure 3b are also true for all orderings, provided $e_0$ is certain (i.e., a constant). Note that if $e_0$ is not a constant, we can introduce a dummy cause $x_l$ that is always instantiated to a nondistinguished value. In this case, we can express the uncertainty in $e_0$ as uncertainty in $e'_{\sigma l}$, leaving $e_0$ a constant in the mathematical formal-ism. Henrion (1987) calls $x_l$ a *base* or *leak* cause for $e$. The following theorem establishes the existence of a set of functions $f_{\sigma i}$ and $g_{\sigma i}$ that satisfy the temporal definition of causal independence when expressed in a diagram such as Figure 3b.

**Theorem 1** *If $e_0$ is a constant, then a set of variables $\{c_1, \ldots, c_n\}$ are causally independent with respect to effect $e$ if and only if for all orderings $\sigma$, there exists function $g_{\sigma 1}$ such that*

$$e_{\sigma 1} \equiv e'_{\sigma 1} = g_{\sigma 1}(c_{\sigma 1}, \epsilon_{\sigma 1}) \quad (2)$$

*and, for $i = 2, \ldots, n$, there exist functions $f_{\sigma i}$ and $g_{\sigma i}$ such that*

$$e_{\sigma i} = f_{\sigma i}(e'_{\sigma i}, e_{\sigma, i-1}) \quad (3)$$
$$e'_{\sigma i} = g_{\sigma i}(c_i, \epsilon_{\sigma i}) \quad (4)$$

**Proof:** The $\Leftarrow$ portion of the theorem follows directly by reading the conditional independence statements associated with causal independence directly from the belief networks associated with each ordering $\sigma$.

We prove $\Rightarrow$ by induction on $n$, the number of causes. When $n = 1$, the theorem follows directly from the transformation described in Figure 3a. For the induction step, let us suppose that $\{c_1, \ldots, c_{n+1}\}$ are causally independent with respect to effect $e$. Let $\sigma$ be the ordering where $c_{\sigma i} = c_i, i = 1, \ldots, n+1$. Applying the theorem to the first $n$ causes, we obtain the belief network in Figure 4a. In particular, we have

$$e = e_{\sigma, n+1} = h_\sigma(c_{n+1}, e_{\sigma n}, \epsilon_{n+1}) \quad (5)$$

for some deterministic function $h_\sigma$. Now, let $\rho$ be the ordering where $c_{\rho 1} = c_{n+1}$ and $c_{\rho, i+1} = c_i, i = 1, \ldots, n$. From the assumption of causal independence, we obtain the belief network in Figure 4b. Specifically, we get

$$e_{\rho 1} = e'_{n+1} = g_{\rho 1}(c_{n+1}, \epsilon_{n+1}) \quad (6)$$

Also, collapsing the functions between $e_{\rho 1} = e'_{n+1}$ and $e_{\rho, n+1}$ in Figure 4b, we obtain

$$e = e_{\rho, n+1} = h_\rho(e'_{n+1}, c_1, \ldots, c_n, \epsilon_1, \ldots, \epsilon_n) \quad (7)$$

for some deterministic function $h_\rho$. Combining Equations 5 and 7, we get

$$\begin{aligned} e &= h_\sigma(c_{n+1}, e_{\sigma n}, \epsilon_{n+1}) \\ &= h_\rho(e'_{n+1}, c_1, \ldots, c_n, \epsilon_1, \ldots, \epsilon_n) \end{aligned} \quad (8)$$

All variables $c_i$ and $\epsilon_i, i = 1, \ldots, n+1$, however, are logically independent (they are also probabilistically independent, but we do not need this fact). Therefore, $e'_{n+1}$ must summarize the effects of $c_{n+1}$ and $\epsilon_{n+1}$ in the determination of $e$, and—similarly—$e_{\sigma n}$ must summarize the effects of $c_1, \ldots, c_n, \epsilon_1, \ldots, \epsilon_n$ in the determination of $e$. Consequently, there must exist some deterministic function $h$ such that

$$e = e_{\sigma, n+1} = e_{\rho, n+1} = h(e'_{n+1}, e_{\sigma n}) \quad (9)$$

Identifying $h$ in Equation 9 with $f_{\sigma, n+1}$ and $g_{\rho 1}$ in Equation 6 with $g_{\sigma, n+1}$, we obtain Equations 2 through

figure=proofa.eps,width=3in (a)
figure=proofb.eps,width=3in (b)

Figure 4: Two belief networks for the proof of Theorem 1.

figure=atemp.eps,width=3in

Figure 5: A belief network representation of a generalized noisy-or model.

3. Repeating this argument for every initial ordering $\sigma$, we complete the induction step. $\square$

An immediate consequence of Theorem 1 is that we can write $e$ as a nested set of two-argument functions for any ordering $\sigma$:

$$e = f_{\sigma n}\left(e'_{\sigma n}, f_{\sigma, n-1}\left(e'_{\sigma, n-1}, \ldots f_{\sigma 2}\left(e'_{\sigma 2}, e'_{\sigma 1}\right)\right)\right) \quad (10)$$

Collapsing these nested functions into a single function $f$, we obtain

$$e = f(e'_{\sigma 1}, \ldots, e'_{\sigma n})$$

We say that Equation 10 is a *nested decomposition* for $f$. In this formulation the sequence of functions that decompose the overall relationship between $e$ and the $e'_{\sigma i}$ depends on the ordering chosen, and each function $f_{\sigma i}$ may be different. If functions $f_{\sigma n}, i = 2, \ldots, n$ are equal to some function $f^*$ for all $\sigma$, however, it follows that the function $f^*$ is both commutative and associative. We find that causal independence relations that are useful in practice have this property, such as the noisy-or and noisy-adder models described in Heckerman (1993).

From our discussion, we see that causal independence is a special case of Srinivas' generalization of the noisy-or model. Srinivas' model is equivalent to the belief network in Figure 5 where the function $f$ is arbitrary. In particular, causal independence includes the assumption that $f$ has a nested decomposition for any ordering of the causes. Indeed, there are many functions $f$ that are not admitted by Heckerman's definition. For example, suppose $e$ is binary. Although both Srinivas' model and causal independence admit a function $f$ that is true if and only if $e'_{\sigma i}$ is true for exactly one value of $i$, only Srinivas' model permits the function $f$ that is true if and only if $e'_{\sigma i}$ is true for exactly two values of $i$, because this function $f$ is not decomposable.

Causal independence also imposes restrictions on the probability distributions for $e'_{\sigma i}$ given $c_{\sigma i}$ and on the individual functions $f_{\sigma i}$. In particular, given the definitions of $e_0$ (a constant) and $e'_{\sigma i}$, it follows that $e'_{\sigma i} = e_0$ when $c_{\sigma i}$ takes on its distinguished value. That is,

$$p(e'_{\sigma i} = e_0 | c_{\sigma i} = *) = 1.0$$

where $*$ is the distinguished value for $c_{\sigma i}$. Furthermore, suppose all variables preceding $c_{\sigma i}$ take on their distinguished value. Then, $e'_{\sigma i} = e_{\sigma i}$. In this situation, however, $e_{\sigma,i-1} = e_0$, by definition of $e_{\sigma,i-1}$. Therefore, from Equation 3, it follows that

$$e_{\sigma i} = f_{\sigma i}(e_{\sigma i}, e_0)$$

That is, $e_0$ is the identity element of each function $f_{\sigma i}$. Thus, for the binary discrete case, the atemporal version of causal independence requires assessment of only $n$ parameters corresponding to $p(e'_{\sigma i} = e_0 | c_{\sigma i} \neq *)$ for a network with all binary variables, as well as the individual functions $f_{\sigma i}$.

It is worth noting that linear models are a special case of causal independence. In a Gaussian linear model we have

$$e = a + \sum_{i=1}^{n} b_i c_i + \varepsilon$$

where $a$ and the $b_i$ are constants and $\varepsilon$ has a normal distribution with mean zero and variance $v$ (written $N(0, v)$). We can express this in terms of the atemporal model by letting

$$e_0 = a$$
$$g_i(c_i, \epsilon_i) = b_i c_i, i = 1, \ldots, n$$
$$g_{n+1}(c_{n+1}, \epsilon_{n+1}) = \epsilon_{n+1}$$
$$\epsilon_{n+1} \sim N(0, v)$$

and by identifying each function $f_{\sigma i}$ with $+$.

## 4 Classes of Causal Interaction

Several types of causal interaction have been proposed in the literature and in this paper. These various classes of causal interaction appear in the following list, ordered from the more general to more specific.

1. **General multiple cause interaction:** Causal interactions modeled with a belief network as shown in Figure 1.

2. **Independence of causal inputs:** Described by Srinivas (1993), and can be modeled in the belief network shown in Figure 5. There is no restriction of the form of $f$.

3. **Singly decomposable causal independence interaction:** There exists some ordering $\sigma$ and set of functions $f_{\sigma i}$, such that Equation 10 holds, as illustrated in Figure 3b.

4. **Fully decomposable causal independence interaction:** Equation 10 and Figure 3b holds for any ordering $\sigma$.

5. **Fully decomposable causal independence with equal functions:** The previous class with the added condition that functions $f_{\sigma n}, i = 2, \ldots, n$ are equal to some function $f^*$ for all $\sigma$, such as *or* or $+$. This class includes the noisy-or model.

6. **Linear Gaussian Models:** A special case of class 5 with continuous-valued causes and effects with a single Gaussian noisy input representing the variance, deterministic contributions for the other causes, and a single function $f^*$ as $+$.

## 5 Causal Independence and Assessment

A major motivation for these prototypical interaction models has been to ease the task of knowledge acquisition for networks where nodes may have many parents. Any formalism at least as specific as that described in class 2 will have economy of knowledge acquisition, since we obtain an exponential savings in parameter assessments. Each more specific class requires even fewer assessments. The appropriateness of each class depends on the particular application. For example, a digital circuit with multiple inputs and a single output can be modeled with a function $f$ as illustrated in Figure 5, but the function $f$ may not be fully or singly decomposable.

Within the class of causal independent models one has a choice of using the temporal or the atemporal definition of causal independence for assessment. The preferred definition depends on the expert and the domain being modeled. For example, in an application involving the effect of drugs on white blood cell counts, the temporal version of causal independence was a more natural method for interacting with the expert [?]. On the other hand, in a number of hardware troubleshooting applications [?], the atemporal version of causal independence (class 5) has been most effective. In these cases, one is typically modeling a device that will fail if any one of it's components fail, leading naturally to a fully decomposable *or* functional model. In general, we find that both definitions are useful in dealing with experts and we can switch between one and the other as needed.

## 6 Causal Independence and Inference

Each more specific class is associated with no worse and usually increased inference-algorithm efficiency. For example, if the function $f$ is decomposable in some ordering—that is as described in class 3—then we can obtain an exponential savings in storage when compared to classes 1 or 2. The effect of the decomposition is to reduce the number of predecessors of the effect node. In addition, if the interactions in a model are causally independent, as in class 4, 5 or 6, then we can rearrange the belief network expression of the decomposition to improve inference further.

As an example of rearranging the belief network, consider the multiply-connected belief network in Figure 6a. If we transform the belief network using the ordering $(c_1, c_2, c_3)$, we obtain the belief network in Figure 6b. In contrast, if we transform the belief net-work using the ordering $(c_2, c_3, c_1)$, then we obtain the belief network in Figure 6c. Inference using exact belief-network algorithms (e.g., junction-tree propagation [?] or arc reversal [?]) may be less efficient in the belief network of Figure 6b than in the belief network of Figure 6c, because there is a larger cycle in the former network. A larger cycle is not necessarily worse for inference in all algorithms, and the desirability of different topologies depends on the specific inference technique being used.

figure=ordera.eps,width=2in (a)
figure=orderb.eps,width=3in (b)
figure=orderc.eps,width=3in (b)

Figure 6: (a) A multiply connected belief network. (b,c) Two equivalent transformations of the belief network in (a). The network in (c) has a smaller undirected cycle than (b).

In the following sections, we quantify savings associated with causal independence due to (1) reducing the size of the predecessor sets and hence clique sizes and (2) rearranging network topologies.

### 6.1 Clique Size Reduction

Although it is clear that there is an exponential savings in storage when using causal independence for a single node, it is less clear what the savings in inference time will be for more general belief networks. Here, we compare state space sizes for networks that use class 1 and class 2 representations of cause-and-effect versus the same models converted to models in class 3 or better. Table 1 shows the size of the maximum clique and the sum of the clique sizes for each network. These figures are proportional to the runtime of clustering style algorithms on belief networks [?].

The BN2 network analyzed in Table 1 is a hypothetical network consisting of ten causes and four effects. Each effect has four causes, and two of the causes are common causes of each effect. With binary nodes, this case shows no savings using causal independence, due to the small state spaces and the small number of parents. In the BN2(5) network, each node was assumed to have 5 outcomes. Here, we obtain a factor of ten savings using causal independence. The Multi-Connected network is an 32 node medical belief network, where most nodes have 2 or 3 states. There is one node in the standard version of the net that has 11 parents. This node and its parents form the largest clique with size 8192 under the standard formulation. Using a class 3 model, the largest clique size becomes 1536; and we obtain a factor of 3 savings in total clique size. The Singly Connected network represents a 27 node hardware diagnosis problem. The network has very few cycles and mostly binary nodes; and there are at most three parents for any causal node. In this case, the additional nodes created in the decomposition result in cliques that were slightly larger than those obtained with the original network. Overall, these results indicate that use of causal independence can have substantial benefits in real-world modeling tasks, especially when a node has many parents and when causes and effects have many outcomes.

figure=graph.eps,width=3in

Figure 7: The frequency distribution of clique sizes for the BN2(5) network under different orderings.

In each of these cases, we used a default ordering for the expansion of the causal independent effect nodes to determine state space size. In the next section, we examine what gains can be expected from searching for the best orderings for expansion, taking into account the overall topology of the graph.

### 6.2 Evaluating Alternate Orderings

Our original hypothesis was that the different orderings of the causal independence expansions of effect nodes could have a large effect on inference. The primary effect of different orderings is to change the size of undirected loops in the original belief network, as illustrated in Figure 6. Under the presumption that large loops are worse than short loops for inference (as has been reported previously), we believed that expansion ordering could have a large effect on clique size and hence inference. It has become apparent, however, that loop size is not a critical determinant, at least for clustering-style algorithms.

In order to characterize the potential savings, we sampled the BN2 style network for different orderings. The distribution of clique sizes for a series of random orderings of causal independence expansions is shown in Figure 7. Note that potential gains are relatively modest: The cliques with the smallest size is only slightly smaller than average. We have developed a search algorithm that combines the process of clipping the diagram with choosing the order of expansion of the causal independence nodes. This algorithm uses heuristics during clique formation to guide the search for good expansions. On the basis of the empirical data, it is likely that a naive ordering will do almost as well.

## 7 Conclusions

In this paper we have developed an atemporal characterization of causal independence. The characterization is based on a functional representation of the interaction between causes and effects that can be written as a nested decomposition of functions. We have shown that when causal independence holds, we easi-ly can covert this decomposition into a belief network that yields efficiency gains in model assessment, storage, and inference.

|  | Classes 1 and 2 | | Class 3 | |
| --- | --- | --- | --- | --- |
| Belief Network | Largest Clique | Sum of Cliques | Largest Clique | Sum of Cliques |
| BN2(binary) | 32 | 128 | 8 | 160 |
| BN2(5) | 3125 | 12500 | 125 | 1250 |
| Multi-Connected | 8192 | 15068 | 1536 | 4966 |
| Singly Connected | 32 | 176 | 32 | 196 |

Table 1: Clique sizes as a function of network and causal-interaction model.

# References

[Druzdel and Simon, 1993] Druzdel, M. and Simon, H. (1993). Causality in Bayesian belief networks. In *Proceedings of Ninth Conference on Uncertainty in Artificial Intelligence,* Washington, DC, pages 3–11. Morgan Kaufmann.

[Henrion, 1987] Henrion, M. (1987). Some practical issues in constructing belief networks. In *Proceedings of the Third Workshop on Uncertainty in Artificial Intelligence,* Seattle, WA, pages 132–139. Association for Uncertainty in Artificial Intelligence, Mountain View, CA. Also in Kanal, L., Levitt, T., and Lemmer, J., editors, *Uncertainty in Artificial Intelligence 3,* pages 161–174. North-Holland, New York, 1989.

[Jensen et al., 1990] Jensen, F., Lauritzen, S., and Olesen, K. (1990). Bayesian updating in recursive graphical models by local computations. *Computational Statisticals Quarterly*, 4:269–282.

[Pearl, 1988] Pearl, J. (1988). *Probabilistic Reasoning in Intelligent Systems: Networks of Plausible Inference.* Morgan Kaufmann, San Mateo, CA.

[Shachter, 1986] Shachter, R. (1986). Evaluating influence diagrams. *Operations Research*, 34:871–882.

[Srinivas, 1993] Srinivas, S. (1993). A generalization of the noisy-Or model. In *Proceedings of Ninth Conference on Uncertainty in Artificial Intelligence,* Washington, DC, pages 208–215. Morgan Kaufmann.